\documentclass[10pt,twocolumn,letterpaper]{article}

\usepackage{iccv}
\usepackage{times}
\usepackage{epsfig}
\usepackage{graphicx}
\usepackage{amsmath}
\usepackage{amssymb}
\usepackage{gensymb}
\usepackage{tabularx}

\usepackage[pagebackref=true,breaklinks=true,letterpaper=true,colorlinks,bookmarks=false]{hyperref}

\iccvfinalcopy 

\def\iccvPaperID{****} 

\ificcvfinal\fi

\newcommand{\secref}[1]{Sec.~\ref{#1}}
\newcommand{\figref}[1]{Fig.~\ref{#1}}
\newcommand{\tabref}[1]{Tab.~\ref{#1}}

\newcommand{\shortname}{X-Mesh}
\newcommand{\modulename}{Text-guided Dynamic Attention Module}
\newcommand{\shortmodulename}{TDAM}
\newcommand{\metricOne}{Multi-view Expert Score}
\newcommand{\metricTwo}{Iteration for Target Score}
\newcommand{\metricOneShort}{MES}
\newcommand{\metricTwoShort}{ITS}
\newcommand{\datasetname}{MIT-30}

\begin{document}

\title{\shortname{}: Towards Fast and Accurate Text-driven 3D Stylization\\ via Dynamic Textual Guidance}

\author{%
  \textbf{Yiwei Ma}$^{1\ddag}$ \quad 
  \textbf{Xiaoqing Zhang}$^{1\ddag}$ \quad 
  \textbf{Xiaoshuai Sun}$^{1}$ 
  \thanks{Corresponding author; $^\ddag$Equal contributions.} \quad  
  \textbf{Jiayi Ji}$^1$ \\ 
  \textbf{Haowei Wang}$^1$ \quad 
  \textbf{Guannan Jiang}$^2$ \quad 
  \textbf{Weilin Zhuang}$^2$ \quad 
  \textbf{Rongrong Ji}$^{1}$
  \\[0.2cm] 
  $^1$Key Laboratory of Multimedia Trusted Perception and Efficient Computing,\\ Ministry of Education of China, Xiamen University, 361005, P.R. China. \\
  $^2$Contemporary Amperex Technology Co., Limited (CATL), Fujian, China  \\[0.152cm]
}

\maketitle
\ificcvfinal\thispagestyle{empty}\fi

\begin{abstract}
Text-driven 3D stylization is a complex and crucial task in the fields of computer vision (CV) and computer graphics (CG), aimed at transforming a bare mesh to fit a target text. Prior methods adopt text-independent multilayer perceptrons (MLPs) to predict the attributes of the target mesh with the supervision of CLIP loss. However, such text-independent architecture lacks textual guidance during predicting attributes, thus leading to \emph{unsatisfactory stylization} and \emph{slow convergence}. To address these limitations, we present \shortname{}, an innovative text-driven 3D stylization framework that incorporates a novel \modulename{} (\shortmodulename{}). The \shortmodulename{} dynamically integrates the guidance of the target text by utilizing text-relevant spatial and channel-wise attentions during vertex feature extraction, resulting in more accurate attribute prediction and faster convergence speed.
Furthermore, existing works lack standard benchmarks and automated metrics for evaluation, often relying on subjective and non-reproducible user studies to assess the quality of stylized 3D assets. To overcome this limitation, we introduce a new standard text-mesh benchmark, namely \datasetname{}, and two automated metrics, which will enable future research to achieve fair and objective comparisons. Our extensive qualitative and quantitative experiments demonstrate that \shortname{} outperforms previous state-of-the-art methods. Our codes and results are available
at our project webpage: \url{https://xmu-xiaoma666.github.io/Projects/X-Mesh/}
\end{abstract}

\begin{figure}[t] 
\centering 
\includegraphics[width=0.50\textwidth]{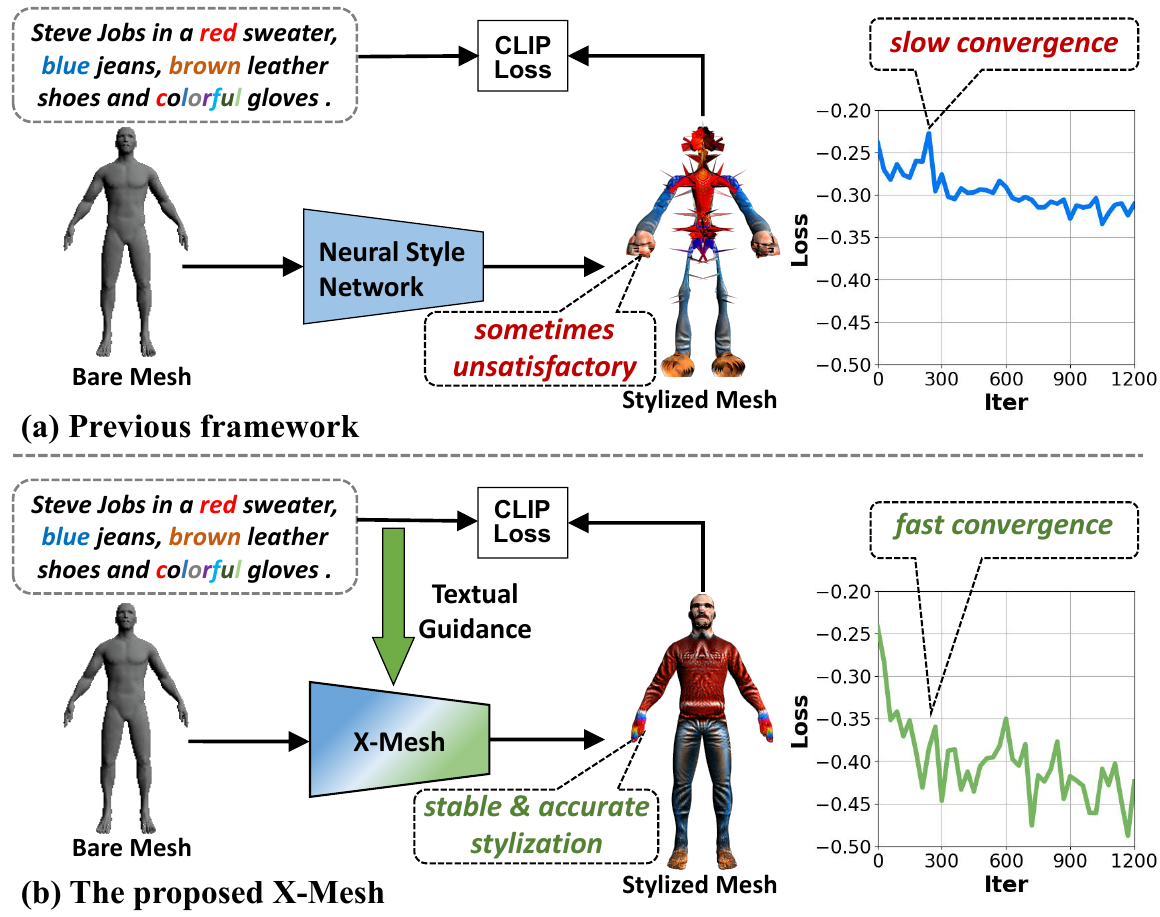} 
\caption{(a) A typical text-driven 3D stylization framework. (b) Our proposed \shortname{} framework. \shortname{} achieves better stylization and faster convergence.} 
\label{fig:intro} 
\end{figure}

\section{Introduction}
In recent years, 3D asset creation through stylization, \emph{i.e.,} transforming bare meshes to match text prompts~\cite{michel2022text2mesh,ChenChenNeurIPS22,wang2022nerf}, images~\cite{wang2022clip,zhang2022arf}, and 3D shapes~\cite{KangxueYin20233DStyleNetC3}, has received significant attention in the fields of computer vision and graphics~\cite{gatys2016image,AaronHertzmann2001ImageA,huang2017arbitrary}.
The resulting stylized 3D assets are applied to a range of practical applications, such as gaming, virtual reality, and film. Among the stylization techniques available, text-driven 3D stylization is particularly user-friendly, as text prompts are more readily available than images or 3D shapes. However, creating stylized 3D assets through text input presents a significant challenge due to the significant gap between visual and linguistic information.

The emergence of Contrastive Language-Image Pre-training (CLIP)~\cite{AlecRadford2021LearningTV} has made it possible to achieve text-driven 3D stylization.
Recently, Text2Mesh~\cite{michel2022text2mesh} and TANGO~\cite{ChenChenNeurIPS22} have made significant contributions in this field by predicting the attributes of each vertex on the mesh with the supervision of CLIP loss. 
Specifically, Text2Mesh predicts the color and displacement of each mesh vertex to generate a stylized mesh that aligns with the target text prompt.
Similarly, TANGO employs neural networks to forecast diffuse, roughness, specular, and normal maps to create photorealistic 3D meshes following a comparable approach.

Despite achieving impressive results, existing text-driven 3D stylization methods have limitations that hinder their effectiveness and efficiency. One major drawback is their failure to fully consider the semantics of the input text during the prediction of mesh vertex attributes.
Current methods only rely on CLIP loss to align the rendered images from the stylized mesh with the text prompt, without any additional textual semantic guidance during predicting vertex attributes. Such approaches lead to several issues, including  \emph{unsatisfactory stylization} and \emph{slow convergence}.
For instance, as shown in \figref{fig:intro}(a), conventional neural style networks do not utilize textual guidance during attribute prediction. As a result, the predicted vertex attributes may not align with the semantic context of the target text prompt, leading to an inconsistent stylized mesh.
Moreover, the lack of additional text guidance makes it difficult to rapidly converge to an acceptable result. Typically, previous methods require over 500 iterations (equivalent to over 8 minutes of training) to attain stable stylized outcomes, which is impractical for users.

To address the issues of inconsistency and slow convergence in conventional neural style networks, we propose \emph{\shortname{}}, a framework that leverages textual semantic guidance to predict vertex attributes. As shown in \figref{fig:intro}(b), \shortname{} produces high-quality stylized results that are consistent with the input text. Besides, with textual guidance during vertex attribute prediction, \shortname{} usually achieves stable results in just 200 iterations (approximately 3 minutes of training).
Our approach  relies on a novel \emph{\modulename{} (\shortmodulename{})} for text-aware attribute prediction. \figref{fig:overview}(b) illustrates how spatial and channel-wise attentions are employed in \shortmodulename{} to extract text-relevant vertex features. Notably, the parameters of the attention modules are dynamically generated by textual features, which makes the vertex features prompt-aware.

Additionally, the quality evaluation of the stylized results from existing text-driven 3D stylization methods~\cite{ChenChenNeurIPS22,michel2022text2mesh} poses a significant challenge. This challenge is mainly reflected in two aspects. 
Firstly, the lack of a standard benchmark for the text-driven 3D stylization problem presents a challenge in evaluating the effectiveness of existing methods. Without fixed text prompts and meshes, the results obtained from previous methods are incomparable. This in turn hinders progress and the development of more effective solutions.
Secondly, the current evaluation of stylized 3D assets relies heavily on user studies, which is a time-consuming and expensive process. Furthermore, this evaluation method is also subject to individual interpretation, which further hinders the reproducibility of results.

To address the aforementioned challenges, we propose a standardized text-mesh benchmark and two automatic evaluation metrics for the fair, objective, and reproducible comparison of text-driven 3D stylization methods. 
The proposed benchmark, called \emph{Mesh wIth Text (\datasetname{})}, contains 30 categories of bare meshes, each of which is annotated with 5 different text prompts for diverse stylization.
The proposed two evaluation metrics aims to overcome the limitations of subjective and non-reproducible user studies used in prior work. 
Specifically, we render 24 images of the stylized 3D mesh from fixed elevation and azimuth angles, and propose two metrics, \emph{\metricOne{} (\metricOneShort{})} and \emph{\metricTwo{} (\metricTwoShort{})}, to evaluate the stylization quality and convergence speed. 

This paper presents two main contributions:

\begin{itemize}
    \item We propose \shortname{} that incorporates a novel text-guided dynamic attention module (TDAM) to improve the accuracy and convergence speed of 3D stylization. 
    \item We construct a standard benchmark and propose two automatic evaluation metrics, which facilitate objective and reproducible assessments of text-driven 3D stylization techniques, and may aid in advancing this field of research.
\end{itemize}

\section{Related Work}
\subsection{Text-to-Image Manipulation/Generation} 
Several previous works have attempted to combine GAN and CLIP to achieve text-to-image generation~\cite{6,53,attngan}. Specifically, StyleGAN~\cite{stylegan1,stylegan2,stylegan3,stylegant} focuses on the latent space to enable better control over generated images. Building on StyleGAN, StyleCLIP~\cite{styleclip} leverages the guidance of CLIP to realize text-to-image generation. DAE-GAN~\cite{daegan} uses a dynamic perception module to comprehensively perceive text information as a development architecture of GAN. Stack-GAN~\cite{stackgan,stackgan++} divides the task into two stages, generating basic color and shape constraints of the objects described in the text and then adding more details to produce high-quality images with high resolution. VQGAN~\cite{vqgan} improves the performance of visual generation on multiple tasks. MirrorGAN~\cite{mirrorgan} combines the global-to-local attention mechanism with a text-to-image-to-text framework to preserve semantics effectively.

Meanwhile, diffusion models have made significant contributions to image generation. DALL-E~\cite{dall-e} and CogView~\cite{cogview,cogview2,cogvideo} are based on transformer and parallel auto-regressive architectures. GLIDE~\cite{glide} leverages classifier-free guidance for image generation and restoration after fine-tuning. DALL-E2~\cite{dall-e2} generates original and realistic images given a text prompt by encoding image features according to the text features of CLIP and then decoding them via a diffusion model. EDiff-I~\cite{ediffi} trains a text-to-image diffusion model for different synthesis stages to achieve high visual quality. Imagen~\cite{imagen} benefits from the semantic encoding ability of the large pre-trained language model T5~\cite{t5} and the diffusion model in generating high-fidelity images.

\subsection{Text-to-3D Manipulation/Generation}

The field of text-to-3D generation has seen significant advancements with the development of text-to-image techniques. Among these techniques, some NeRF-based methods have shown promise, especially when used in combination with CLIP. Some notable examples of such methods include CLIP-NeRF~\cite{clip-nerf}, PureCLIPNeRF~\cite{pureclipnerf}, and DreamFields~\cite{dreamfields}. Additionally, recent studies have explored the fusion of CLIP with other algorithms, such as ISS~\cite{iss} with SVR~\cite{svr}, CLIP-Forge~\cite{clip-forge} using a normalizing flow network~\cite{flow-network}, and AvatarCLIP~\cite{avatarclip} leveraging SMLP~\cite{smpl}.
Furthermore, the diffusion model~\cite{diffusion} has recently demonstrated impressive results in text-to-image generation, leading to its integration into the text-to-3D generation process. Examples of studies that have incorporated the diffusion model into their generation process include DreamFusion~\cite{dreamfusion}, Magic3D~\cite{lin2022magic3d}, and Dream3D~\cite{xu2022dream3d}.

Besides, mesh-based stylization is also widely researched due to its wide applicability.
Traditionally, the stylization of bare meshes in computer graphics requires professional knowledge. However, recent studies~\cite{siddiqui2022texturify,gao2022get3d} have made strides in the automation of stylizing 3D representations using text prompts. For instance, CLIP-Mesh~\cite{mohammad2022clip-mesh} uses CLIP and loop subdivision~\cite{loopsubdivision} to achieve 3D asset generation. While TANGO~\cite{ChenChenNeurIPS22} incorporates reflection knowledge, it is limited in shape manipulation. Text2Mesh~\cite{michel2022text2mesh}, on the other hand, predicts both color and displacement of each vertex to achieve stronger stylization.
This paper proposes a text-guided dynamic attention module in the vertex attribute prediction phase. This module not only leads to a better stylization effect but also achieves a fast convergence speed.

\subsection{Attention Mechanism}
Attention mechanism is a widely-used technique in deep learning that has been applied to a variety of tasks, including computer vision~\cite{hu2018squeeze, wang2023towards, hu2021istr, hu2023you}, natural language processing~\cite{luong2015effective, shang2015neural, vaswani2017attention}, and multimodal fields~\cite{ma2023towards,xu2015show, ma2022knowing, ma2022x, ye2022shifting,ji2022knowing}.
The concept of attention was first introduced in the context of neural machine translation by Bahdanau \emph{et al.}~\cite{bahdanau2014neural}, who proposed a model that learns to align the source and target sentences by focusing on different parts of the source sentence at each decoding step. Since then, various attention mechanisms have been proposed to improve the performance of different models.
For example, Hu \emph{et al.}~\cite{hu2018squeeze} proposed channel attention to enhance the image recognition ability of the model. Woo \emph{et al.}~\cite{woo2018cbam} leveraged both channel attention and spatial attention to focus on important areas and channels. Ye \emph{et al.}~\cite{ye2022shifting} introduced dynamic attention for visual grounding, where different visual features are generated for different referring expressions. Self-attention~\cite{vaswani2017attention}, which is an effective global attention mechanism first proposed for NLP tasks, has been widely used to improve the performance of different models. Wang \emph{et al.}~\cite{wang2018non} introduced a non-local attention mechanism for video understanding tasks. Liu \emph{et al.}~\cite{liu2021swin} improved self-attention by introducing shifted windows, which enhances the local perception ability of the model.
In this paper, we propose a text-guided dynamic attention mechanism for text-driven 3D stylization, which enables the spatial (vertex) and channel information of the input mesh to be dynamically focused based on the target text prompt.

\section{Approach}

In this section, we first explain the overall architecture of \shortname{} in \secref{sec:32}. Then, we provide the details of the proposed \modulename{} in \secref{sec:33}.

\begin{figure*}[t] 
\centering 
\includegraphics[width=1.00\textwidth]{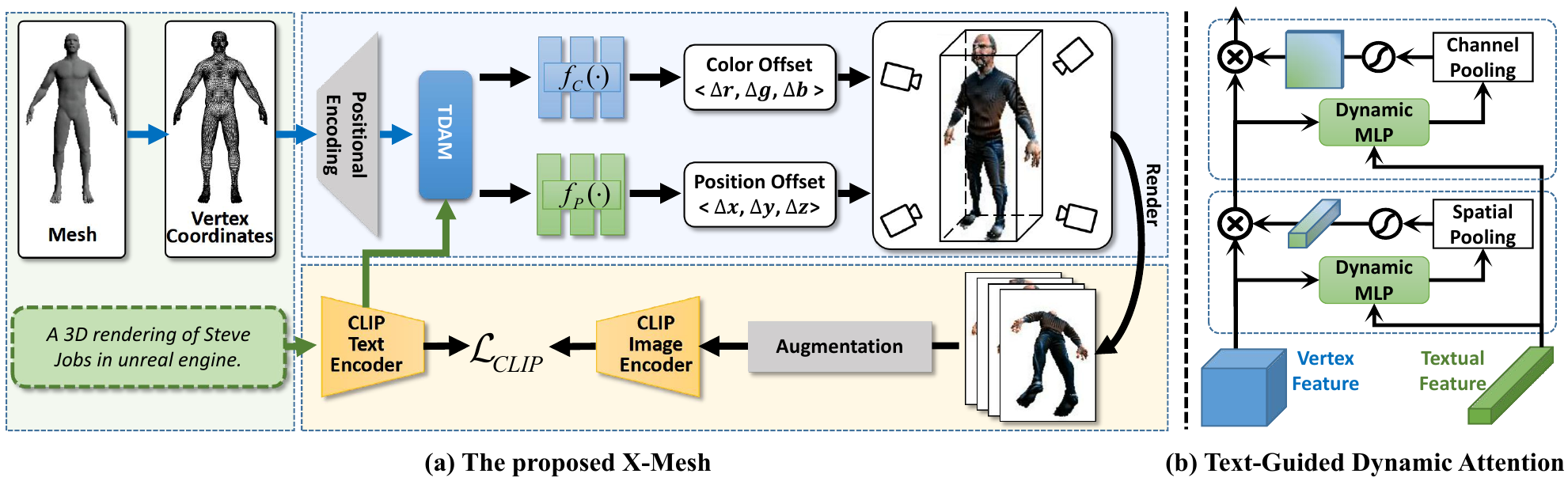} 
\caption{(a) Illustration of the proposed \shortname{} model, which modifies the appearance and geometry of the input mesh according to the text prompt. 
(b) An overview of \shortmodulename{}, which aims to process vertex features under the guidance of target text.} 
\label{fig:overview} 
\end{figure*}

\subsection{Architecture} \label{sec:32}

An illustration of the proposed \shortname{} is shown in \figref{fig:overview}(a). The goal of \shortname{} is to modify an input mesh to match a given text prompt by predicting its appearance and geometry. Specifically, an input mesh $\mathcal{M}$ is defined as a set of vertices $\mathcal{V} \in \mathbb{R}^{n \times 3}$ and faces $\mathcal{F} \in \{1, \ldots, n\}^{m \times 3}$, which are kept constant during training. Here, $n$ and $m$ denote the number of vertices and faces, respectively. Given an input mesh and a target text prompt, \shortname{} predicts the appearance attribute (\emph{i.e.,} the color offset $\Delta C_p \in \mathbb{R}^{3}$) and the geometry attribute (\emph{i.e.,} the position offset $\Delta P_p \in \mathbb{R}^{3}$) of each vertex $p \in \mathcal{V}$, and finally generates a stylized mesh $\mathcal{M^S}$ that conforms to the target text.

We start by initializing the color of each vertex to $(0.5, 0.5, 0.5)$ and normalizing the vertex coordinates to fit within a unit cube. To synthesize high-frequency details, we apply positional encoding using Fourier feature mappings to each vertex. Specifically, given a vertex $p \in \mathcal{V}$ of the mesh, we compute the positional encoding $PE_{(p)}$ as follows:
\begin{equation}
PE_{(p)}=[\cos (2 \pi \mathbf{B} p), \sin (2 \pi \mathbf{B} p)]^{\mathrm{T}},
\label{eq:pe}
\end{equation}
where $\mathbf{B} \in \mathbb{R}^{C \times 3}$ is a random Gaussian matrix, and each value in this matrix is randomly sampled from a normal distribution with mean 0 and variance $\sigma^2$.

Then, the proposed \shortmodulename{} takes in the vertex positional encoding feature $PE_{(p)}$, which is dynamically processed under the guidance of the target text prompt. The resulting feature is further passed through two MLP branches, the Color MLP $f_C(\cdot)$ and the Position MLP $f_P(\cdot)$, which generate the color offset $\Delta C_p$ and the position offset $\Delta P_p$, respectively. Following \cite{michel2022text2mesh}, the position offset $\Delta P_p$ is constrained to a small value, specifically $\left| \Delta P_p \right|_2 \leq 0.1$, to prevent excessive deformation.
The new color and position attributes of each point are defined as $C_p^\prime = C_p + \Delta C_p$ and $P_p^\prime = P_p + \Delta P_p$, respectively. Here, $C_p \in \mathbb{R}^3$ and $P_p \in \mathbb{R}^3$ represent the RGB color and coordinates of $p$ on the original input mesh, respectively.
To enhance geometry, a gray stylized mesh $\mathcal{M}^{\mathcal{S}}_{gray}$ is used, which has the same geometry as $\mathcal{M}^{S}$ but the color of all vertices are set to gray.

We employ an interpolation-based differentiable renderer~\cite{chen2019learning} for $\mathcal{M}^\mathcal{S}$ and $\mathcal{M}^\mathcal{S}_{gray}$ from $n_\theta$ different views. For each view $\theta$, we could obtain two rendered images, \emph{i.e.,} $I_\theta^{color}$ for $\mathcal{M}^\mathcal{S}$ and $I_\theta^{gray}$ for $\mathcal{M}^\mathcal{S}_{gray}$. We then apply 2D augmentation $\psi(\cdot)$ to each rendered image, and extract their features using the CLIP visual encoder $E_v(\cdot)$~\cite{AlecRadford2021LearningTV}.
We obtain the final feature representation by averaging the features across all views, which can be formulated as follows:
\begin{equation} 
    \phi_{color} = \frac{1}{n_\theta}\sum_\theta  E_v(\psi(I_\theta^{color})),
\label{eq:color}
\end{equation}
\begin{equation} 
    \phi_{gray} = \frac{1}{n_\theta}\sum_\theta  E_v(\psi(I_\theta^{gray})),
\label{eq:gray}
\end{equation}

To align the rendered images and the target text in CLIP space, we adopt CLIP textual encoder $E_t(\cdot)$ to embed the text prompt. The framework is trained using the CLIP loss, and the training objective can be formulated as:
\begin{equation} 
    \mathcal{L} = - \text{sim}\left(\phi_{color},E_t(\mathcal{T})\right)-\text{sim}\left(\phi_{gray},E_t(\mathcal{T})\right),
\label{eq:loss}
\end{equation}
where $\mathcal{T}$ represents the target text prompt, and $\text{sim}(a,b)$ denotes the cosine similarity between $a$ and $b$.

\subsection{\modulename{}} \label{sec:33}
Previous works on text-driven 3D stylization have been limited by their inability to fully exploit the target text to guide the prediction of vertex attributes, resulting in suboptimal stylization results. To address this limitation, we propose a novel \modulename{} (\shortmodulename{}) that leverages the target text to guide the attribute prediction process. An overview of our approach is shown in \figref{fig:overview}(b), which illustrates how \shortmodulename{} calculates text-related vertex attention at both channel and spatial levels. Our proposed \shortmodulename{} is based on a dynamic linear layer, whose parameters are generated dynamically based on the target textual features. We first explain how the dynamic linear layer is implemented and then describe how we design \shortmodulename{} based on this layer to compute text-aware dynamic channel and spatial attention maps.

\noindent \textbf{Dynamic Linear Layer.}
Existing text-driven 3D stylization methods use static MLPs to predict the attributes of each vertex on the mesh. However, since the parameters of these MLPs are randomly generated, the target text cannot provide additional guidance during attribute prediction. 
To address this limitation, we propose a dynamic linear layer, whose parameters are generated based on the target textual feature  $\mathbf{F}_t \in \mathbb{R}^{D_t}$. The dynamic linear layer is defined as follows:
\begin{equation} 
    \mathbf{x}_{out} = \mathbf{x}_{in} \mathbf{W}_t +\mathbf{b}_t,
\label{eq:dynamic}
\end{equation}
where $\mathbf{x}_{in} \in \mathbb{R}^{D_{in}}$ and $\mathbf{x}_{out} \in \mathbb{R}^{D_{out}}$ represent the input and output vectors of the dynamic linear layer, respectively. 
The trainable parameters of the dynamic linear layer are denoted as $\mathbf{M}_d \in \mathbb{R}^{(D_{in}+1) \times D_{out}} =\{\mathbf{W}_t \in \mathbb{R}^{D_{in} \times D_{out}}, \mathbf{b}_t \in \mathbb{R}^{D_{out}}\} $, which are generated based on the target textual feature $\mathbf{F}_t$. 

A straightforward method to generate dynamic parameters is to use a plain linear layer, defined as follows:
\begin{equation} 
    \mathbf{M}_d = \mathbf{F}_t \mathbf{W}_{m} + \mathbf{b}_{m},
\label{eq:linear}
\end{equation}
where $\mathbf{W}_{m} \in \mathbb{R}^{D_{t} \times (D_{in}+1)*D_{out}}$ and $\mathbf{b}_{m} \in \mathbb{R}^{(D_{in}+1)*D_{out}}$. However, this method requires a large number of trainable parameters, specifically ${(D_{t}+1) * (D_{in}+1)*D_{out}}$, which can result in an unaffordable training cost and overfitting.

Thus, we use matrix decomposition to reduce the number of trainable parameters. Specifically, We decompose $\mathbf{M}_d  \in \mathbb{R}^{(D_{in}+1) \times D_{out}}$ into $\mathbf{U} \in \mathbb{R}^{(D_{in}+1) \times K}$ and $\mathbf{V} \in \mathbb{R}^{K \times D_{out}}$, where $K$ is a hyper-parameter that determines the compression ratio. It can be formulated as follows:
\begin{equation} 
    \mathbf{M}_d = \mathbf{U} \mathbf{V},
\label{eq:uv}
\end{equation}
where $\mathbf{U}$ is a parameter matrix dynamically generated from $\mathbf{F}_t$ and $\mathbf{V}$ is a static trainable matrix. The formulation of $\mathbf{U}$ is presented as follows:
\begin{equation} 
    \mathbf{U} = \Phi(\mathbf{F}_t \mathbf{W}_l + \mathbf{b}_l),
\label{eq:u}
\end{equation}
where $\mathbf{W}_l \in \mathbb{R}^{D_{t} \times (D_{in}+1)*K}$ and $\mathbf{b}_l \in \mathbb{R}^{(D_{in}+1)*K}$. $\Phi(\cdot)$ is a reshape function that transfers the input from $\mathbb{R}^{(D_{in}+1)*K}$ to $\mathbb{R}^{(D_{in}+1) \times K}$.

Through the matrix decomposition technique, the number of trainable parameters is reduced from ${(D_{t}+1) \times (D_{in}+1) * D_{out}}$ to $(D_{t}+1) \times (D_{in}+1) * K + K \times D_{out}$, which saves on additional training cost and avoids the risk of over-fitting.

\noindent \textbf{Dynamic Channel and Spatial Attention.}
As explained earlier, our goal is to obtain vertex features that are sensitive to the target text. To achieve this, we propose a \modulename{} (\shortmodulename{}) that builds upon the dynamic linear layer and comprises two types of attention mechanisms, \emph{i.e.,} channel attention and spatial attention.

The key element of \shortmodulename{} is the dynamic MLP, which comprises two dynamic linear layers separated by a ReLU activation function. Inspired by squeeze-and-excitation networks~\cite{hu2018squeeze}, the input and output dimensions of the dynamic MLP are identical, while the hidden dimension is reduced by a factor $r$.

In \shortmodulename{}, the objective of channel attention is to activate the channels of the vertex feature that are related to the target text.
Specifically, given the vertex feature $\mathbf{F}_v \in \mathbb{R}^{N_v \times D_v}$, where $N_v$ is the number of vertices of the input mesh and $D_v$ is the channel dimension of the input mesh, we first pass it through a dynamic MLP and then aggregate spatial dimensions through average pooling. To obtain the channel-wise attention map, we normalize the values to a range of 0 to 1 using the Sigmoid activation function as follows:
\begin{equation} 
    \mathbf{A}_{ca} = \sigma\left(\frac{1}{N_v}\sum_{i=1}^{N_v}\eta_1(\mathbf{F}_v)[i,:]\right),
\label{eq:mlp_ca}
\end{equation}
where $\mathbf{A}_{ca} \in \mathbb{R}^{1 \times D_v}$ denotes the channel-wise attention map, $\sigma(\cdot)$ represents the Sigmoid function, and $\eta_1(\cdot)$ refers to the dynamic MLP. To obtain the channel-activated vertex feature $\mathbf{F}_v^{\prime} \in \mathbb{R}^{N_v \times D_v}$, we take the element-wise product of $\mathbf{F}_v$ and $\mathbf{A}_{ca}$ as follows:
\begin{equation} 
    \mathbf{F}_v^{\prime} = \mathbf{F}_v \otimes \mathbf{A}_{ca},
\label{eq:channel}
\end{equation}
where $\otimes$ is the element-wise product.

The goal of spatial attention in \shortmodulename{} is to activate the vertices that are related to the target text. First, we feed the channel-activated vertex feature $\mathbf{F}^\prime$ into another dynamic MLP and aggregate the channel dimensions using the average function. The output is then normalized using the Sigmoid activation function as follows:
\begin{equation} 
    \mathbf{A}_{sa} = \sigma\left(\frac{1}{D_v}\sum_{j=1}^{D_v}\eta_2(\mathbf{F}^\prime_v)[:,j]\right),
\label{eq:mlp_sa}
\end{equation}
where $\mathbf{A}_{sa} \in \mathbb{R}^{N_v \times 1}$, and $\eta_2(\cdot)$ is a dynamic MLP with non-shared parameters with $\eta_1(\cdot)$.
Finally, to obtain the spatially-activated vertex feature $\mathbf{F}_v^{\prime \prime}$, we perform element-wise product between $\mathbf{F}_v^\prime$ and $\mathbf{A}_{sa}$:
\begin{equation} 
    \mathbf{F}_v^{\prime \prime} = \mathbf{F}^\prime_v \otimes \mathbf{A}_{sa}.
\label{eq:spatial}
\end{equation}

\section{Benchmarks and Metrics}

\noindent \textbf{Benchmark.}
In this paper, we construct a text-mesh benchmark to standardize the evaluation process of text-driven 3D stylization. The proposed \datasetname{} benchmark includes 30 categories of bare meshes, collected from various public 3D datasets such as COSEG~\cite{sidi2011unsupervised}, Thingi10K~\cite{zhou2016thingi10k}, Shapenet~\cite{chang2015shapenet}, Turbo Squid~\cite{turbosquid}, and ModelNet~\cite{wu20153d}. To ensure a diverse range of stylization, each mesh is annotated with five different text prompts. We found that the prompt template of \emph{`A 3D rendering of $\cdots$ in unreal engine.'} is a good default, so all meshes are annotated with this prompt template if not specified.

\begin{figure*}[t] 
\centering 
\includegraphics[width=1.00\textwidth]{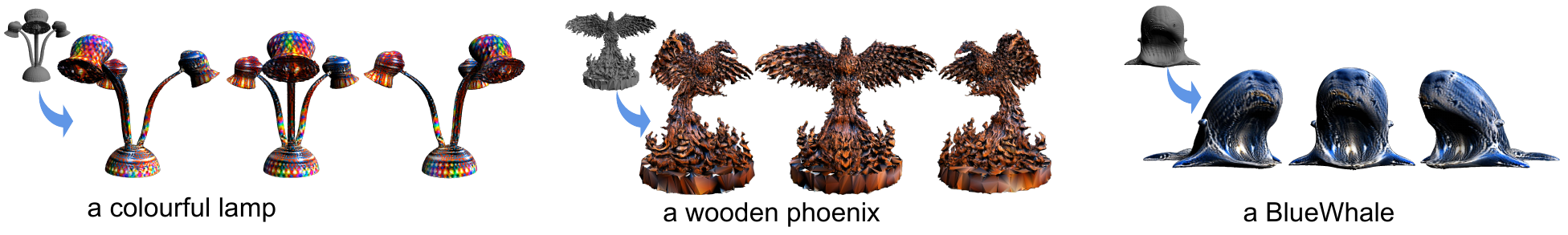} 
\caption{Text-driven 3D stylization results. \shortname{} provides high-quality stylization results for a collection of prompts and meshes.} 
\label{fig:result} 
\end{figure*}

\begin{figure}[t] 
\centering 
\includegraphics[width=0.46\textwidth]{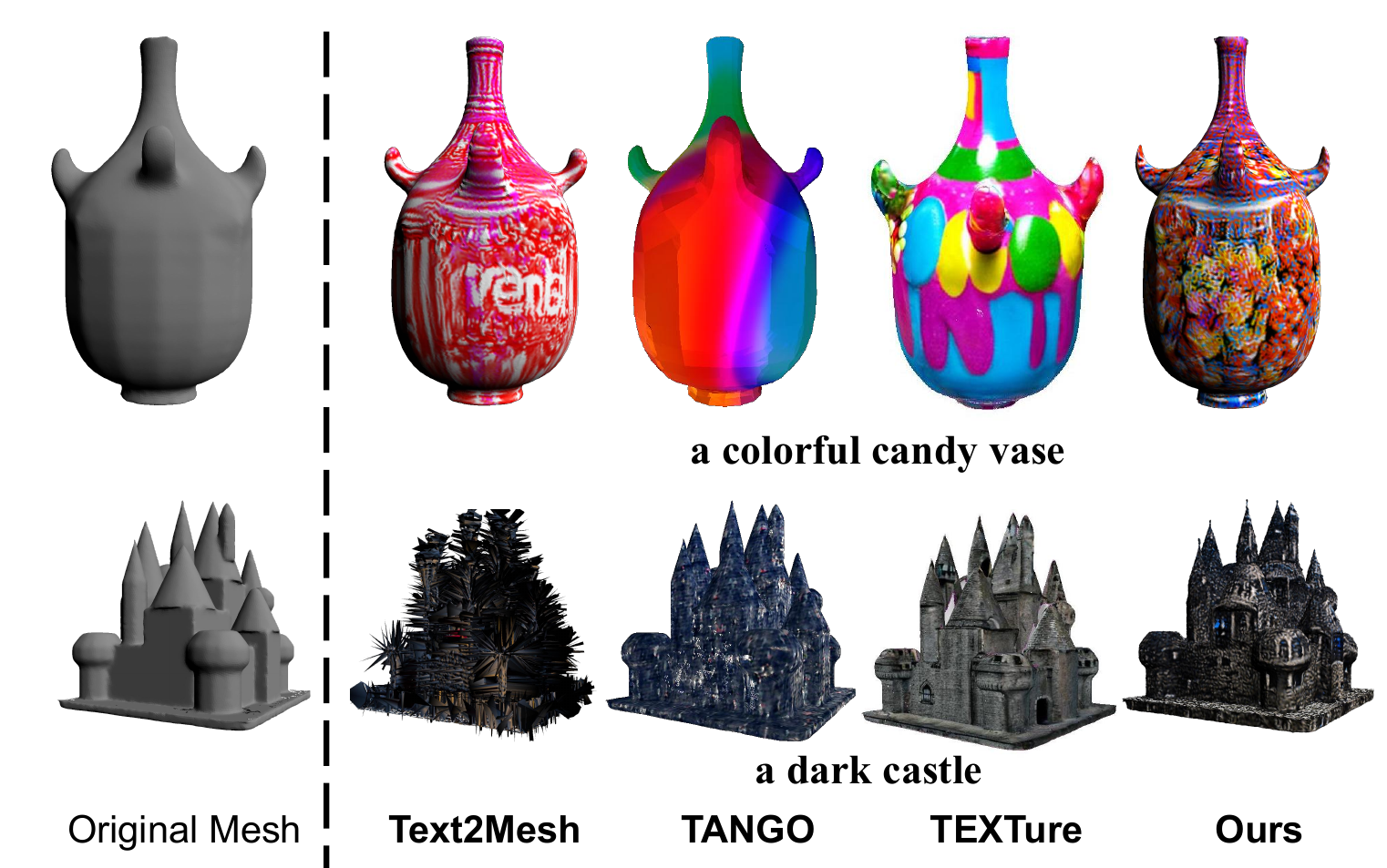} 
\caption{Text-driven 3D stylization results of Text2Mesh~\cite{michel2022text2mesh}, TANGO~\cite{ChenChenNeurIPS22}, TEXTure~\cite{richardson2023texture}, and \shortname{} (Ours) given the same mesh and prompt. \shortname{} provides high-quality and realistic stylization results.} 
\label{fig:compare} 
\end{figure}

\noindent \textbf{Metrics.} %
Some previous works~\cite{ChenChenNeurIPS22,michel2022text2mesh} have used user studies to evaluate the perceived quality of stylized 3D assets, which is often subjective and non-reproducible. Other works~\cite{jain2022zero,lee2022understanding} have employed the metric~\cite{park2021benchmark} for text-to-image generation to assess the quality of 3D assets. However, this metric does not account for the continuity of 3D assets, as it only measures the similarity between a single-angle rendered image of the 3D asset and the target text. Given that text-driven 3D stylization aims to produce a 3D asset that conforms to the target text, evaluating rendered images from multiple angles is necessary.

To enable objective and reproducible comparisons, we propose two automatic metrics that are based on multi-angle rendered images of 3D assets. These metrics will replace manual evaluation in user studies, allowing for a reliable evaluation of text-driven 3D stylization methods.

Given a stylized 3D asset, we begin by rendering 24 images $\mathbf{I} = \{I_i\}_{i=1}^{24}$ from 24 fixed views, taking into account both azimuth angle $\theta_\text{azi}$ and elevation angle $\theta_\text{ele}$. For each 3D asset, we establish a standard view where $\theta_\text{azi} = 0\degree$ and $\theta_\text{ele} = 0\degree$. Using this standard view as a basis, we leverage 8 azimuth angles (0\degree, 45\degree, 90\degree, 135\degree, 180\degree, 225\degree, 270\degree, 315\degree) and 3 elevation angles (-30\degree, 0\degree, 30\degree) to render 24 rendered images. To address the subjective and non-reproducible nature of user studies, we use an automatic expert model~\cite{cherti2022reproducible}~\footnote{\url{https://github.com/mlfoundations/open_clip}} trained on LAION-400M~\cite{schuhmann2021laion} for evaluation. Based on these 24 rendered images and the expert model, we propose two automatic evaluation metrics. Specifically, \metricOneShort{} is used to evaluate the extent to which the stylized 3D asset conforms to the target text, and \metricTwoShort{} is used to evaluate the convergence rate of the model.

 \begin{figure*}[t] 
\centering 
\includegraphics[width=0.85\textwidth]{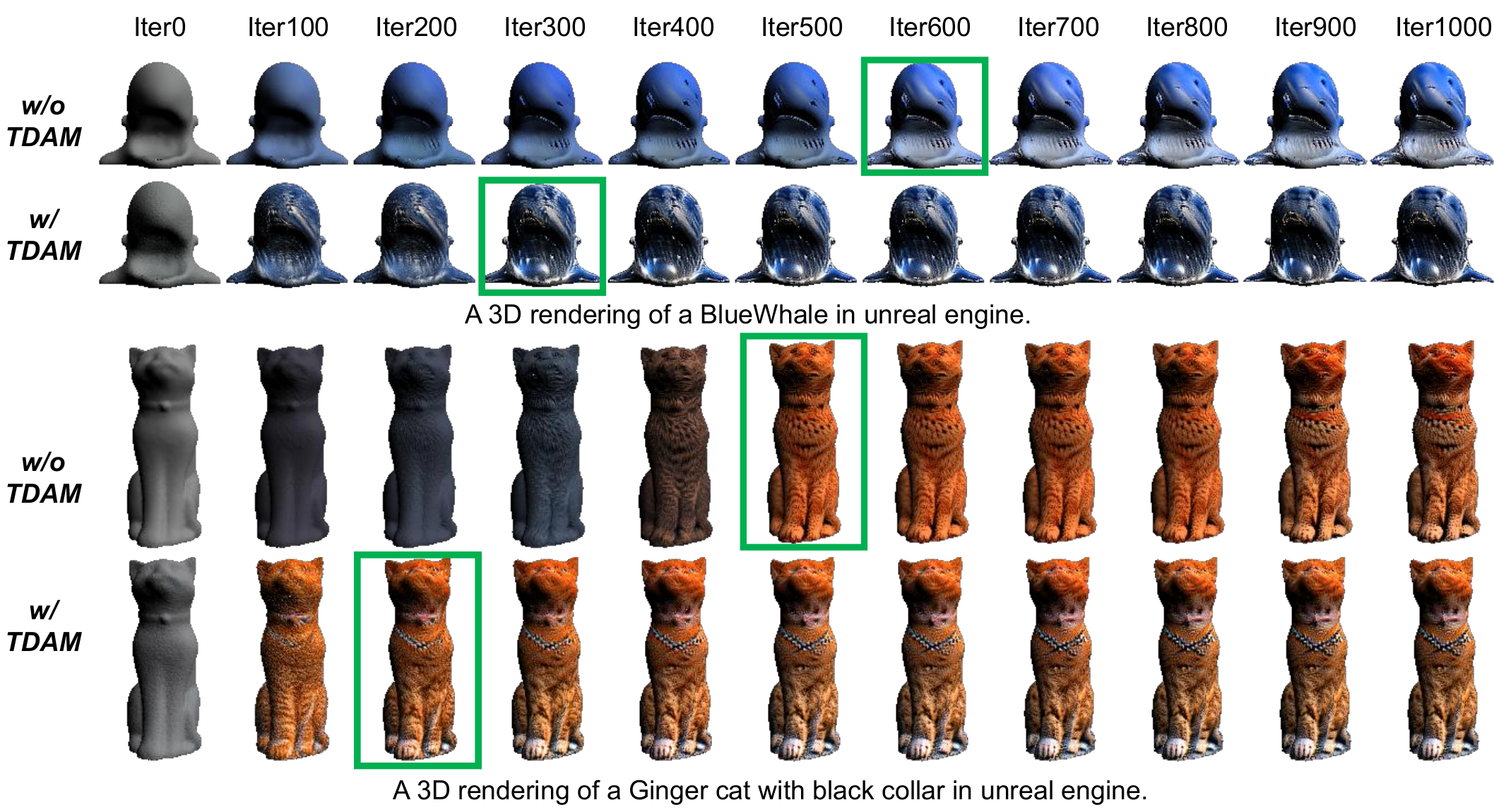} 
\caption{Visualization of text-driven 3D stylization process with and without the proposed \shortmodulename{} under different iterations. The green box indicates the  first iteration to obtain a stable stylization result.} 
\label{fig:convergence} 
\end{figure*}

For \metricOneShort{}, we first embed the 24 rendered images and the corresponding text prompt into a shared space using the visual and textual encoders of the expert model. Then, we calculate the cosine similarity scores between the rendered images and the corresponding text, and obtain \metricOneShort{} by averaging them. The formulation of \metricOneShort{} is as follows:
\begin{equation}
\text{\metricOneShort{}}(\mathcal{M}^\mathcal{S},\mathcal{T}) = \frac{1}{24} \sum_{i=1}^{24} \text{sim}\Big(E^\prime_v(I_i),E^\prime_t(\mathcal{T})\Big),
\label{eq:metric1}
\end{equation}
where $\mathcal{M}^\mathcal{S}$ and $\mathcal{T}$ is the stylized 3D mesh and the corresponding text prompt, respectively. $E^\prime_v(\cdot)$ and $E^\prime_t(\cdot)$ refer to the visual encoder and textual encoder of the expert model.

\metricTwoShort{} represents the minimum number of iterations needed to achieve the target \metricOneShort{}. For instance, $\text{\metricTwoShort{}}_{0.3}(\mathcal{M}^\mathcal{S},\mathcal{T})$ indicates the minimum number of iterations required when $\text{\metricOneShort{}}(\mathcal{M}^\mathcal{S},\mathcal{T})=0.3$. In our experiment, we set the maximum number of training iterations for each mesh to 1200. If a mesh fails to reach the target \metricOneShort{} within 1200 iterations, we set  \metricTwoShort{} of this sample to 2000. The final \metricOneShort{} and \metricTwoShort{} are obtained by averaging them across all samples in the benchmark.

\section{Experiments}

We conducted all experiments using the public PyTorch library on a single RTX 3090 24GB GPU. We trained our proposed \shortname{} using the Adam optimizer with a learning rate of $5e\text{-}4$. We set $C$, $n_\theta$, $r$, $\sigma$, and $K$ to 256, 5, 8, 12, and 30, respectively. $\psi(\cdot)$ includes RandomPerspective and RandomResizedCrop.  Our method typically achieves high-quality stylized results in just 3 minutes due to its fast convergence rate. In comparison, previous methods~\cite{ChenChenNeurIPS22,michel2022text2mesh} typically take more than 8 minutes to produce stable results on the same GPU.

In \secref{sec:compare}, we qualitatively compare \shortname{} with state-of-the-art text-driven 3D stylization approaches on \datasetname{}. In \secref{sec:ablation}, we conduct the ablation study to explore the effectiveness of the proposed module. Finally, We evaluate our method and previous SOTA methods with quantitative metrics in \secref{sec:quantitative}. 

\subsection{Text-driven Stylization}\label{sec:compare}

\noindent \textbf{Qualitative Results.} %
\figref{fig:result} showcases some stylized results generated by \shortname{} for various meshes and driving prompts. The results demonstrate that the stylized meshes are not only faithful to the target text, but also visually plausible. For instance, when given the prompt \emph{``a colourful lamp''}, \shortname{} produces a lamp with vibrant colors that match the prompt while preserving the lamp's shape and structure. Moreover, the generated outputs exhibit a high degree of consistency across different viewpoints. For instance, when given the prompt \emph{``a wooden phoenix''}, the rendered images from different angles exhibit consistent stylization.

\begin{figure*}[t] 
\centering 
\includegraphics[width=1.00\textwidth]{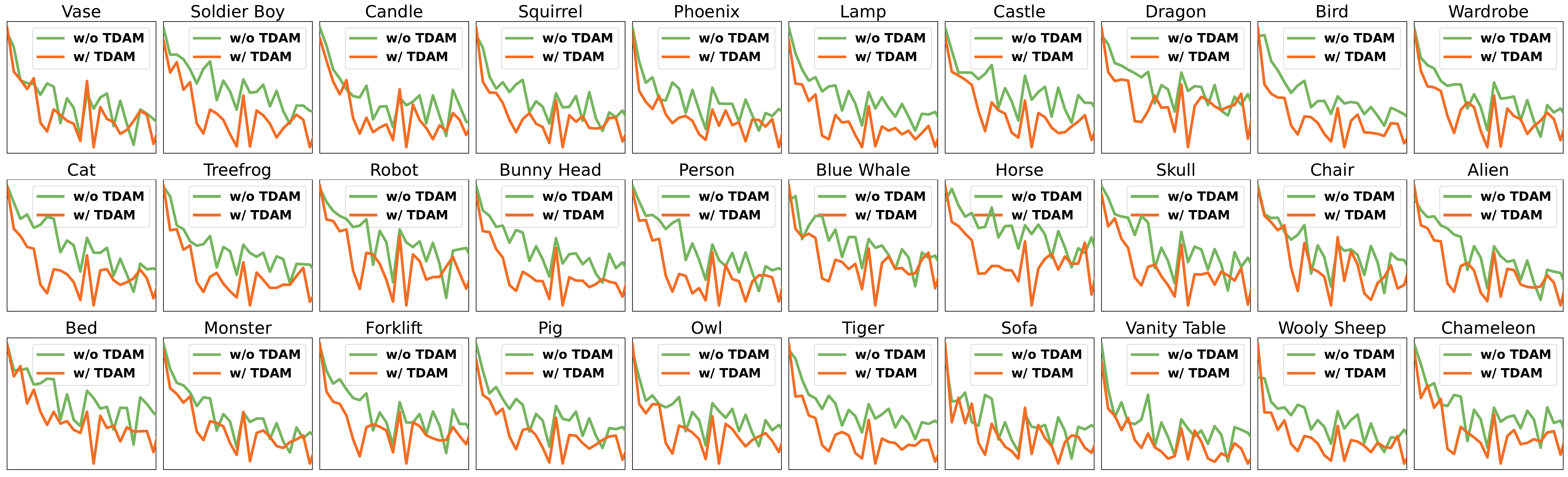} 
\caption{The loss change of each mesh category during training for models with and without \shortmodulename{}, where the loss values of 5 prompts for each mesh are averaged. The x-axis represents the training iteration, and the y-axis is the loss value. Due to the limitation of page space, we have omitted the contents of the x-axis and y-axis. See \emph{supplementary materials} for a detailed version.} 
\label{fig:losses} 
\end{figure*}

\begin{figure}[t] 
\centering 
\includegraphics[width=0.50\textwidth]{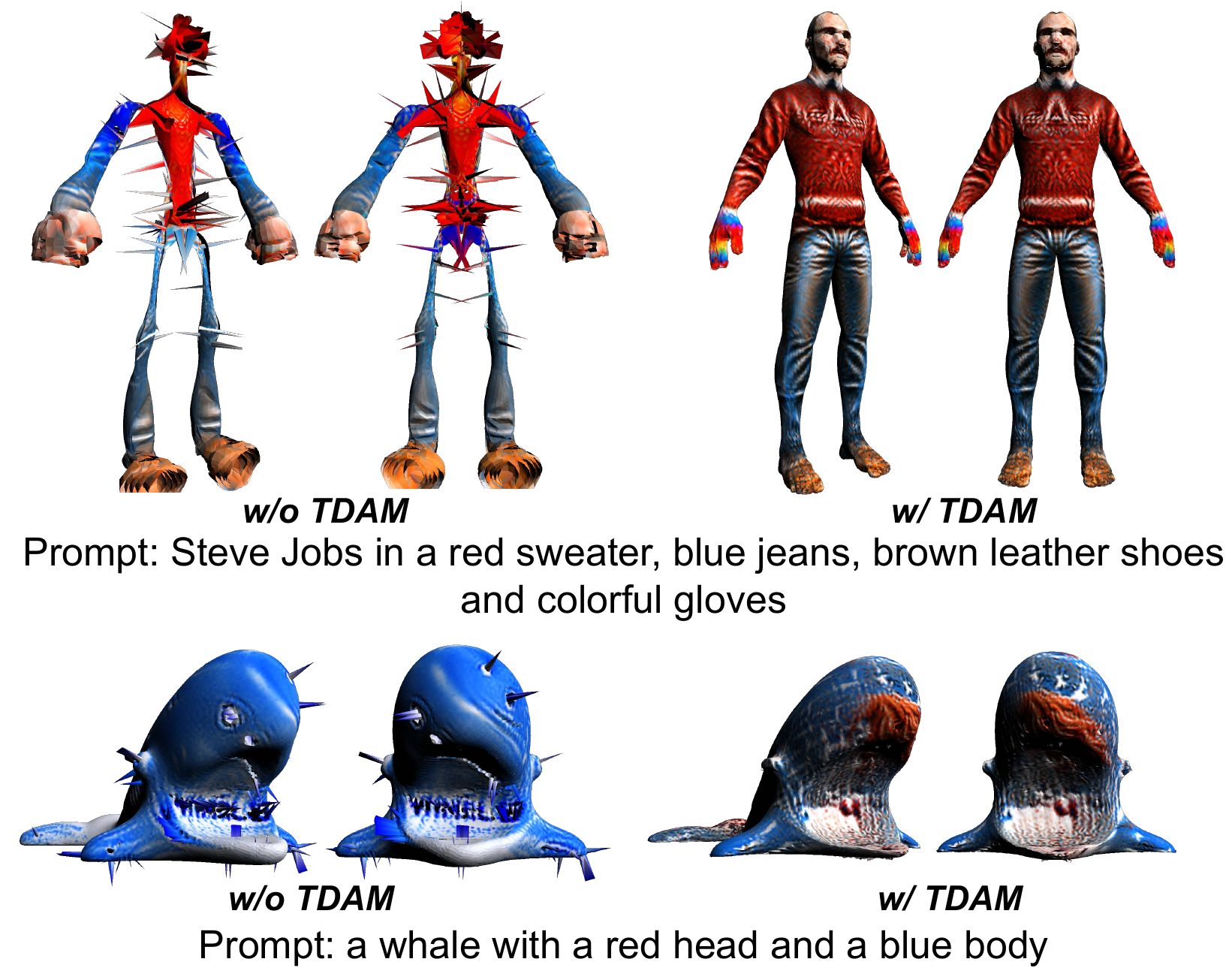} 
\caption{Qualitative comparison of 3D assets generated based on complex prompts without \shortmodulename{} and with \shortmodulename{}.} 
\label{fig:complex} 
\end{figure}

\noindent \textbf{Qualitative Comparisons.} %
In this comparison study presented in \figref{fig:compare}, we provide evidence of the superiority of our proposed method, \shortname{}, over several existing state-of-the-art approaches for text-driven 3D stylization. We observe that Text2Mesh~\cite{michel2022text2mesh} frequently produces unreasonable deformation, which can be attributed to excessive displacement of vertices. For instance, we provide the example of the \emph{``a dark castle''} in the bottom part of \figref{fig:compare}, where Text2Mesh generates several spikes that do not conform to the original structure of the castle.

On the other hand, TANGO~\cite{ChenChenNeurIPS22} and TEXTure~\cite{richardson2023texture}, which do not displace the vertices of the original mesh, do not suffer from the deformation problem observed in Text2Mesh. However, They still has several shortcomings in terms of stylization quality and text understanding. We demonstrate this by showing the example of the \emph{``a colorful candy vase''} in the top part of \figref{fig:compare}, where TANGO and TEXTure simply apply several colors to the vase without taking into account its underlying structure.

In contrast, our proposed method, \shortname{}, overcomes both issues and generates textures that conform to the target text through proper displacement and color prediction for each vertex. We attribute this advantage to the introduction of dynamic guidance of text during vertex attribute prediction. By incorporating dynamic textual guidance, our method is able to generate more accurate results that are in line with the target text.

\subsection{Ablation Study}\label{sec:ablation}

\noindent \textbf{Convergence Speed.} 
Convergence speed is a crucial factor to consider when assessing the effectiveness of text-driven 3D stylization. A fast-converging model enables users to obtain the desired 3D asset for a given prompt quickly, while a slow-converging model can be frustrating for users to wait for. The results presented in \figref{fig:convergence} demonstrate that our proposed \shortmodulename{} significantly improves convergence speed, allowing the model to reach an acceptable result in under 100 iterations. In contrast, the model without \shortmodulename{} requires more than 300 iterations to achieve similar results. Additionally, the \shortmodulename{}-equipped model reaches stable results in fewer than 300 iterations, while the model without \shortmodulename{} requires more than 500 iterations. This remarkable improvement in convergence speed can be attributed to the \shortmodulename{} module, which introduces textual guidance in the attribute prediction process. As a result, the proposed model achieves faster convergence speeds, making it an efficient and effective solution for text-driven 3D stylization.

Moreover, in \figref{fig:losses}, we provide the loss curves for 30 categories of meshes in \datasetname{}. These curves illustrate that the loss value of the model with \shortmodulename{} decreases faster than that of the model without \shortmodulename{} during training. The superior performance of the proposed model suggests that \shortmodulename{} significantly improves the efficiency and effectiveness of text-driven 3D stylization, making it a highly promising tool for 3D content creation. Overall, these findings underscore the importance of \shortmodulename{} for text-driven 3D stylization, which can significantly improve the convergence speed and reduce the training time for users.

\noindent \textbf{Robustness to Complex Prompts.} %
In this section, we aim to investigate the ability of the proposed \shortname{} to handle complex text prompts with the aid of the \shortmodulename{} module. To achieve this goal, we conduct several experiments with complex prompts and report our observations as follows:

Firstly, we observe that the model without the \shortmodulename{} module is highly susceptible to collapse when presented with complex prompts. In particular, as shown in the first line of \figref{fig:complex}, the final stylized mesh exhibits numerous spikes and loses its normal geometry when the model lacks the \shortmodulename{} module. In contrast, the \shortmodulename{}-equipped model can accurately predict the appropriate color and geometric attributes that match the target text.

Furthermore, we observe that the model without \shortmodulename{} may fail to capture some critical details in complex prompts. For example, the model without \shortmodulename{} ignores \emph{``black collar''} in the third line of \figref{fig:convergence} and \emph{``colorful gloves''} in the first line of \figref{fig:complex}. By contrast, our method can make accurate predictions through comprehensive text understanding.

Overall, our experimental results demonstrate that the \shortmodulename{}-enhanced model can effectively handle complex text prompts and produce high-quality stylized 3D meshes.

\subsection{Quantitative Comparison}\label{sec:quantitative}

\begin{table}
\centering
\begin{tabularx}{0.5\textwidth}{>{\raggedright\arraybackslash}X|>{\centering\arraybackslash}X>{\centering\arraybackslash}X}
        \hline
        & \metricOneShort{} $\uparrow$ & \metricTwoShort{}$_{0.22}$ $\downarrow$ \\ \hline
        TANGO~\cite{ChenChenNeurIPS22} & 23.21 & 795.47\\
        Text2Mesh~\cite{michel2022text2mesh} & 28.85 & 173.27 \\
        X-Mesh & \textbf{29.26} & \textbf{88.53}  \\ \hline
\end{tabularx}
\caption{Qualitative comparison of state-of-the-art methods for text-driven 3D stylization. Note that a higher \metricOneShort{} and a lower \metricTwoShort{}$_{0.22}$ is preferable in this table.}
\label{tab:quantitative}
\end{table}

In previous works, user studies are used to evaluate stylization results. However, this evaluation approach has limitations, as it is subjective and non-reproducible. To overcome these limitations, we propose two automatic evaluation metrics, \metricOneShort{} and \metricTwoShort{}, which respectively measure the quality of the stylized assets and the convergence speed of stylization models.
As presented in \tabref{tab:quantitative}, our proposed \shortname{} outperforms previous methods. Specifically, \shortname{} achieves a 0.41 absolute improvement in \metricOneShort{} on \datasetname{}, indicating that our method produces better stylization quality than previous works. Moreover, \shortname{} obtains the lowest \metricTwoShort{}$_{0.22}$, highlighting that our method converges faster than previous methods. The superior performance of our proposed method demonstrated in both \metricOneShort{} and \metricTwoShort{}, metrics further validates the effectiveness and superiority of \shortname{} over previous methods, and supports its potential for practical applications.

\section{Conclusion}
In this paper, we propose \shortname{}, a novel text-driven 3D stylization framework that leverages a text-guided dynamic attention module to predict vertex attributes, resulting in accurate stylization and fast convergence. 
Furthermore, we construct a text-mesh benchmark and introduce two automatic metrics to facilitate an objective and reproducible evaluation of this field. Extensive experiments demonstrate that \shortname{} outperforms existing state-of-the-art methods both qualitatively and quantitatively. 

\section*{Acknowledgement}
This work was supported by National Key R\&D Program of China (No.2022ZD0118201), the National Science Fund for Distinguished Young Scholars (No.62025603), the National Natural Science Foundation of China (No. U21B2037, No. U22B2051, No. 62176222, No. 62176223, No. 62176226, No. 62072386, No. 62072387, No. 62072389, No. 62002305 and No. 62272401), China Postdoctoral Science Foundation (No.2023M732948), and the Natural Science Foundation of Fujian Province of China (No.2021J01002,  No.2022J06001).

{\small
\bibliographystyle{ieee_fullname}
\bibliography{egbib}

\begin{thebibliography}{10}\itemsep=-1pt

\bibitem{bahdanau2014neural}
Dzmitry Bahdanau, Kyunghyun Cho, and Yoshua Bengio.
\newblock Neural machine translation by jointly learning to align and
  translate.
\newblock {\em arXiv preprint arXiv:1409.0473}, 2014.

\bibitem{ediffi}
Yogesh Balaji, Seungjun Nah, Xun Huang, Arash Vahdat, Jiaming Song, Karsten
  Kreis, Miika Aittala, Timo Aila, Samuli Laine, Bryan Catanzaro, Tero Karras,
  and Ming-Yu Liu.
\newblock ediffi: Text-to-image diffusion models with an ensemble of expert
  denoisers.
\newblock 2022.

\bibitem{chang2015shapenet}
Angel~X Chang, Thomas Funkhouser, Leonidas Guibas, Pat Hanrahan, Qixing Huang,
  Zimo Li, Silvio Savarese, Manolis Savva, Shuran Song, Hao Su, et~al.
\newblock Shapenet: An information-rich 3d model repository.
\newblock {\em arXiv preprint arXiv:1512.03012}, 2015.

\bibitem{6}
Hila Chefer, Sagie Benaim, Roni Paiss, and Lior Wolf.
\newblock Image-based clip-guided essence transfer.
\newblock In {\em Computer Vision--ECCV 2022: 17th European Conference, Tel
  Aviv, Israel, October 23--27, 2022, Proceedings, Part XIII}, pages 695--711.
  Springer, 2022.

\bibitem{chen2019learning}
Wenzheng Chen, Huan Ling, Jun Gao, Edward Smith, Jaakko Lehtinen, Alec
  Jacobson, and Sanja Fidler.
\newblock Learning to predict 3d objects with an interpolation-based
  differentiable renderer.
\newblock {\em Advances in neural information processing systems}, 32, 2019.

\bibitem{ChenChenNeurIPS22}
Yongwei Chen, Rui Chen, Jiabao Lei, Yabin Zhang, and Kui Jia.
\newblock Tango: Text-driven photorealistic and robust 3d stylization via
  lighting decomposition.
\newblock In {\em Proceedings of the Neural Information Processing Systems
  (NeurIPS)}, 2022.

\bibitem{cherti2022reproducible}
Mehdi Cherti, Romain Beaumont, Ross Wightman, Mitchell Wortsman, Gabriel
  Ilharco, Cade Gordon, Christoph Schuhmann, Ludwig Schmidt, and Jenia Jitsev.
\newblock Reproducible scaling laws for contrastive language-image learning.
\newblock {\em arXiv preprint arXiv:2212.07143}, 2022.

\bibitem{cogview}
Ming Ding, Zhuoyi Yang, Wenyi Hong, Wendi Zheng, Chang Zhou, Da Yin, Junyang
  Lin, Xu Zou, Zhou Shao, Hongxia Yang, et~al.
\newblock Cogview: Mastering text-to-image generation via transformers.
\newblock {\em Advances in Neural Information Processing Systems},
  34:19822--19835, 2021.

\bibitem{cogview2}
Ming Ding, Wendi Zheng, Wenyi Hong, and Jie Tang.
\newblock Cogview2: Faster and better text-to-image generation via hierarchical
  transformers.
\newblock {\em arXiv preprint arXiv:2204.14217}, 2022.

\bibitem{flow-network}
Laurent Dinh, Jascha Sohl-Dickstein, and Samy Bengio.
\newblock Density estimation using real nvp.
\newblock {\em arXiv preprint arXiv:1605.08803}, 2016.

\bibitem{vqgan}
Patrick Esser, Robin Rombach, and Bj{\"o}rn Ommer.
\newblock Taming transformers for high-resolution image synthesis.
\newblock 2020.

\bibitem{glide}
Richard~A Friesner, Jay~L Banks, Robert~B Murphy, Thomas~A Halgren, Jasna~J
  Klicic, Daniel~T Mainz, Matthew~P Repasky, Eric~H Knoll, Mee Shelley, Jason~K
  Perry, et~al.
\newblock Glide: a new approach for rapid, accurate docking and scoring. 1.
  method and assessment of docking accuracy.
\newblock {\em Journal of medicinal chemistry}, 47(7):1739--1749, 2004.

\bibitem{gao2022get3d}
Jun Gao, Tianchang Shen, Zian Wang, Wenzheng Chen, Kangxue Yin, Daiqing Li, Or
  Litany, Zan Gojcic, and Sanja Fidler.
\newblock Get3d: A generative model of high quality 3d textured shapes learned
  from images.
\newblock {\em Advances In Neural Information Processing Systems},
  35:31841--31854, 2022.

\bibitem{gatys2016image}
Leon~A Gatys, Alexander~S Ecker, and Matthias Bethge.
\newblock Image style transfer using convolutional neural networks.
\newblock In {\em Proceedings of the IEEE conference on computer vision and
  pattern recognition}, pages 2414--2423, 2016.

\bibitem{AaronHertzmann2001ImageA}
Aaron Hertzmann, Charles~E. Jacobs, Nuria Oliver, Brian Curless, and David
  Salesin.
\newblock Image analogies.
\newblock {\em international conference on computer graphics and interactive
  techniques}, 2001.

\bibitem{avatarclip}
Fangzhou Hong, Mingyuan Zhang, Liang Pan, Zhongang Cai, Lei Yang, and Ziwei
  Liu.
\newblock Avatarclip: Zero-shot text-driven generation and animation of 3d
  avatars.
\newblock {\em arXiv preprint arXiv:2205.08535}, 2022.

\bibitem{cogvideo}
Wenyi Hong, Ming Ding, Wendi Zheng, Xinghan Liu, and Jie Tang.
\newblock Cogvideo: Large-scale pretraining for text-to-video generation via
  transformers.
\newblock {\em arXiv preprint arXiv:2205.15868}, 2022.

\bibitem{hu2021istr}
Jie Hu, Liujuan Cao, Yao Lu, ShengChuan Zhang, Yan Wang, Ke Li, Feiyue Huang,
  Ling Shao, and Rongrong Ji.
\newblock Istr: End-to-end instance segmentation with transformers.
\newblock {\em arXiv preprint arXiv:2105.00637}, 2021.

\bibitem{hu2023you}
Jie Hu, Linyan Huang, Tianhe Ren, Shengchuan Zhang, Rongrong Ji, and Liujuan
  Cao.
\newblock You only segment once: Towards real-time panoptic segmentation.
\newblock In {\em Proceedings of the IEEE/CVF Conference on Computer Vision and
  Pattern Recognition}, pages 17819--17829, 2023.

\bibitem{hu2018squeeze}
Jie Hu, Li Shen, and Gang Sun.
\newblock Squeeze-and-excitation networks.
\newblock In {\em Proceedings of the IEEE conference on computer vision and
  pattern recognition}, pages 7132--7141, 2018.

\bibitem{huang2017arbitrary}
Xun Huang and Serge Belongie.
\newblock Arbitrary style transfer in real-time with adaptive instance
  normalization.
\newblock In {\em Proceedings of the IEEE international conference on computer
  vision}, pages 1501--1510, 2017.

\bibitem{dreamfields}
Ajay Jain, Ben Mildenhall, Jonathan~T Barron, Pieter Abbeel, and Ben Poole.
\newblock Zero-shot text-guided object generation with dream fields.
\newblock In {\em Proceedings of the IEEE/CVF Conference on Computer Vision and
  Pattern Recognition}, pages 867--876, 2022.

\bibitem{jain2022zero}
Ajay Jain, Ben Mildenhall, Jonathan~T Barron, Pieter Abbeel, and Ben Poole.
\newblock Zero-shot text-guided object generation with dream fields.
\newblock In {\em Proceedings of the IEEE/CVF Conference on Computer Vision and
  Pattern Recognition}, pages 867--876, 2022.

\bibitem{ji2022knowing}
Jiayi Ji, Yiwei Ma, Xiaoshuai Sun, Yiyi Zhou, Yongjian Wu, and Rongrong Ji.
\newblock Knowing what to learn: a metric-oriented focal mechanism for image
  captioning.
\newblock {\em IEEE Transactions on Image Processing}, 31:4321--4335, 2022.

\bibitem{stylegan3}
Tero Karras, Miika Aittala, Samuli Laine, Erik H{\"a}rk{\"o}nen, Janne
  Hellsten, Jaakko Lehtinen, and Timo Aila.
\newblock Alias-free generative adversarial networks.
\newblock {\em Advances in Neural Information Processing Systems}, 34:852--863,
  2021.

\bibitem{stylegan1}
Tero Karras, Samuli Laine, and Timo Aila.
\newblock A style-based generator architecture for generative adversarial
  networks.
\newblock In {\em Proceedings of the IEEE/CVF conference on computer vision and
  pattern recognition}, pages 4401--4410, 2019.

\bibitem{stylegan2}
Tero Karras, Samuli Laine, Miika Aittala, Janne Hellsten, Jaakko Lehtinen, and
  Timo Aila.
\newblock Analyzing and improving the image quality of stylegan.
\newblock In {\em Proceedings of the IEEE/CVF conference on computer vision and
  pattern recognition}, pages 8110--8119, 2020.

\bibitem{pureclipnerf}
Han-Hung Lee and Angel~X Chang.
\newblock Understanding pure clip guidance for voxel grid nerf models.
\newblock {\em arXiv preprint arXiv:2209.15172}, 2022.

\bibitem{lee2022understanding}
Han-Hung Lee and Angel~X Chang.
\newblock Understanding pure clip guidance for voxel grid nerf models.
\newblock {\em arXiv preprint arXiv:2209.15172}, 2022.

\bibitem{lin2022magic3d}
Chen-Hsuan Lin, Jun Gao, Luming Tang, Towaki Takikawa, Xiaohui Zeng, Xun Huang,
  Karsten Kreis, Sanja Fidler, Ming-Yu Liu, and Tsung-Yi Lin.
\newblock Magic3d: High-resolution text-to-3d content creation.
\newblock {\em arXiv preprint arXiv:2211.10440}, 2022.

\bibitem{iss}
Zhengzhe Liu, Peng Dai, Ruihui Li, Xiaojuan Qi, and Chi-Wing Fu.
\newblock Iss: Image as stetting stone for text-guided 3d shape generation.
\newblock {\em arXiv preprint arXiv:2209.04145}, 2022.

\bibitem{liu2021swin}
Ze Liu, Yutong Lin, Yue Cao, Han Hu, Yixuan Wei, Zheng Zhang, Stephen Lin, and
  Baining Guo.
\newblock Swin transformer: Hierarchical vision transformer using shifted
  windows.
\newblock In {\em Proceedings of the IEEE/CVF international conference on
  computer vision}, pages 10012--10022, 2021.

\bibitem{loopsubdivision}
Charles Loop.
\newblock Smooth subdivision surfaces based on triangles.
\newblock 1987.

\bibitem{smpl}
Matthew Loper, Naureen Mahmood, Javier Romero, Gerard Pons-Moll, and Michael~J
  Black.
\newblock Smpl: A skinned multi-person linear model.
\newblock {\em ACM transactions on graphics (TOG)}, 34(6):1--16, 2015.

\bibitem{luong2015effective}
Minh-Thang Luong, Hieu Pham, and Christopher~D Manning.
\newblock Effective approaches to attention-based neural machine translation.
\newblock {\em arXiv preprint arXiv:1508.04025}, 2015.

\bibitem{ma2023towards}
Yiwei Ma, Jiayi Ji, Xiaoshuai Sun, Yiyi Zhou, and Rongrong Ji.
\newblock Towards local visual modeling for image captioning.
\newblock {\em Pattern Recognition}, 138:109420, 2023.

\bibitem{ma2022knowing}
Yiwei Ma, Jiayi Ji, Xiaoshuai Sun, Yiyi Zhou, Yongjian Wu, Feiyue Huang, and
  Rongrong Ji.
\newblock Knowing what it is: semantic-enhanced dual attention transformer.
\newblock {\em IEEE Transactions on Multimedia}, 2022.

\bibitem{ma2022x}
Yiwei Ma, Guohai Xu, Xiaoshuai Sun, Ming Yan, Ji Zhang, and Rongrong Ji.
\newblock X-clip: End-to-end multi-grained contrastive learning for video-text
  retrieval.
\newblock In {\em Proceedings of the 30th ACM International Conference on
  Multimedia}, pages 638--647, 2022.

\bibitem{michel2022text2mesh}
Oscar Michel, Roi Bar-On, Richard Liu, Sagie Benaim, and Rana Hanocka.
\newblock Text2mesh: Text-driven neural stylization for meshes.
\newblock In {\em Proceedings of the IEEE/CVF Conference on Computer Vision and
  Pattern Recognition}, pages 13492--13502, 2022.

\bibitem{mohammad2022clip-mesh}
Nasir Mohammad~Khalid, Tianhao Xie, Eugene Belilovsky, and Tiberiu Popa.
\newblock Clip-mesh: Generating textured meshes from text using pretrained
  image-text models.
\newblock In {\em SIGGRAPH Asia 2022 Conference Papers}, pages 1--8, 2022.

\bibitem{svr}
Michael Niemeyer, Lars Mescheder, Michael Oechsle, and Andreas Geiger.
\newblock Differentiable volumetric rendering: Learning implicit 3d
  representations without 3d supervision.
\newblock In {\em Proceedings of the IEEE/CVF Conference on Computer Vision and
  Pattern Recognition}, pages 3504--3515, 2020.

\bibitem{park2021benchmark}
Dong~Huk Park, Samaneh Azadi, Xihui Liu, Trevor Darrell, and Anna Rohrbach.
\newblock Benchmark for compositional text-to-image synthesis.
\newblock In {\em Thirty-fifth Conference on Neural Information Processing
  Systems Datasets and Benchmarks Track (Round 1)}, 2021.

\bibitem{styleclip}
Or Patashnik, Zongze Wu, Eli Shechtman, Daniel Cohen-Or, and Dani Lischinski.
\newblock Styleclip: Text-driven manipulation of stylegan imagery.
\newblock In {\em Proceedings of the IEEE/CVF International Conference on
  Computer Vision}, pages 2085--2094, 2021.

\bibitem{dreamfusion}
Ben Poole, Ajay Jain, Jonathan~T Barron, and Ben Mildenhall.
\newblock Dreamfusion: Text-to-3d using 2d diffusion.
\newblock {\em arXiv preprint arXiv:2209.14988}, 2022.

\bibitem{mirrorgan}
Tingting Qiao, Jing Zhang, Duanqing Xu, and Dacheng Tao.
\newblock Mirrorgan: Learning text-to-image generation by redescription.
\newblock In {\em Proceedings of the IEEE/CVF Conference on Computer Vision and
  Pattern Recognition}, pages 1505--1514, 2019.

\bibitem{AlecRadford2021LearningTV}
Alec Radford, Jong~Wook Kim, Chris Hallacy, Aditya Ramesh, Gabriel Goh,
  Sandhini Agarwal, Girish Sastry, Amanda Askell, Pamela Mishkin, Jack Clark,
  Gretchen Krueger, and Ilya Sutskever.
\newblock Learning transferable visual models from natural language
  supervision.
\newblock {\em international conference on machine learning}, 2021.

\bibitem{t5}
Colin Raffel, Noam Shazeer, Adam Roberts, Katherine Lee, Sharan Narang, Michael
  Matena, Yanqi Zhou, Wei Li, and Peter~J Liu.
\newblock Exploring the limits of transfer learning with a unified text-to-text
  transformer.
\newblock {\em The Journal of Machine Learning Research}, 21(1):5485--5551,
  2020.

\bibitem{dall-e2}
Aditya Ramesh, Prafulla Dhariwal, Alex Nichol, Casey Chu, and Mark Chen.
\newblock Hierarchical text-conditional image generation with clip latents.
\newblock 2023.

\bibitem{dall-e}
Aditya Ramesh, Mikhail Pavlov, Gabriel Goh, Scott Gray, Chelsea Voss, Alec
  Radford, Mark Chen, and Ilya Sutskever.
\newblock Zero-shot text-to-image generation.
\newblock In {\em International Conference on Machine Learning}, pages
  8821--8831. PMLR, 2021.

\bibitem{53}
Scott Reed, Zeynep Akata, Xinchen Yan, Lajanugen Logeswaran, Bernt Schiele, and
  Honglak Lee.
\newblock Generative adversarial text to image synthesis.
\newblock In {\em International conference on machine learning}, pages
  1060--1069. PMLR, 2016.

\bibitem{richardson2023texture}
Elad Richardson, Gal Metzer, Yuval Alaluf, Raja Giryes, and Daniel Cohen-Or.
\newblock Texture: Text-guided texturing of 3d shapes.
\newblock {\em arXiv preprint arXiv:2302.01721}, 2023.

\bibitem{daegan}
Shulan Ruan, Yong Zhang, Kun Zhang, Yanbo Fan, Fan Tang, Qi Liu, and Enhong
  Chen.
\newblock Dae-gan: Dynamic aspect-aware gan for text-to-image synthesis.
\newblock {\em arXiv: Computer Vision and Pattern Recognition}, 2021.

\bibitem{diffusion}
Chitwan Saharia, William Chan, Saurabh Saxena, Lala Li, Jay Whang, Emily
  Denton, Seyed Kamyar~Seyed Ghasemipour, Burcu~Karagol Ayan, S~Sara Mahdavi,
  Rapha~Gontijo Lopes, et~al.
\newblock Photorealistic text-to-image diffusion models with deep language
  understanding.
\newblock {\em arXiv preprint arXiv:2205.11487}, 2022.

\bibitem{imagen}
Chitwan Saharia, William Chan, Saurabh Saxena, Lala Li, Jay Whang, Emily
  Denton, Seyed Kamyar, Seyed Ghasemipour, Burcu Karagol, S~Sara Mahdavi,
  Rapha~Gontijo Lopes, Tim Salimans, Jonathan Ho, David~J Fleet, and Mohammad
  Norouzi.
\newblock Photorealistic text-to-image diffusion models with deep language
  understanding.
\newblock 2023.

\bibitem{clip-forge}
Aditya Sanghi, Hang Chu, Joseph~G Lambourne, Ye Wang, Chin-Yi Cheng, Marco
  Fumero, and Kamal~Rahimi Malekshan.
\newblock Clip-forge: Towards zero-shot text-to-shape generation.
\newblock In {\em Proceedings of the IEEE/CVF Conference on Computer Vision and
  Pattern Recognition}, pages 18603--18613, 2022.

\bibitem{stylegant}
Axel Sauer, Tero Karras, Samuli Laine, Andreas Geiger, and Timo Aila.
\newblock Stylegan-t: Unlocking the power of gans for fast large-scale
  text-to-image synthesis.
\newblock {\em arXiv preprint arXiv:2301.09515}, 2023.

\bibitem{schuhmann2021laion}
Christoph Schuhmann, Richard Vencu, Romain Beaumont, Robert Kaczmarczyk,
  Clayton Mullis, Aarush Katta, Theo Coombes, Jenia Jitsev, and Aran
  Komatsuzaki.
\newblock Laion-400m: Open dataset of clip-filtered 400 million image-text
  pairs.
\newblock {\em arXiv preprint arXiv:2111.02114}, 2021.

\bibitem{shang2015neural}
Lifeng Shang, Zhengdong Lu, and Hang Li.
\newblock Neural responding machine for short-text conversation.
\newblock {\em arXiv preprint arXiv:1503.02364}, 2015.

\bibitem{siddiqui2022texturify}
Yawar Siddiqui, Justus Thies, Fangchang Ma, Qi Shan, Matthias Nie{\ss}ner, and
  Angela Dai.
\newblock Texturify: Generating textures on 3d shape surfaces.
\newblock In {\em European Conference on Computer Vision}, pages 72--88.
  Springer, 2022.

\bibitem{sidi2011unsupervised}
Oana Sidi, Oliver van Kaick, Yanir Kleiman, Hao Zhang, and Daniel Cohen-Or.
\newblock Unsupervised co-segmentation of a set of shapes via descriptor-space
  spectral clustering.
\newblock In {\em Proceedings of the 2011 SIGGRAPH Asia Conference}, pages
  1--10, 2011.

\bibitem{turbosquid}
TurboSquid.
\newblock Turbosquid 3d model repository, 2021.
\newblock https://www.turbosquid.com/.

\bibitem{vaswani2017attention}
Ashish Vaswani, Noam Shazeer, Niki Parmar, Jakob Uszkoreit, Llion Jones,
  Aidan~N Gomez, {\L}ukasz Kaiser, and Illia Polosukhin.
\newblock Attention is all you need.
\newblock {\em Advances in neural information processing systems}, 30, 2017.

\bibitem{wang2022clip}
Can Wang, Menglei Chai, Mingming He, Dongdong Chen, and Jing Liao.
\newblock Clip-nerf: Text-and-image driven manipulation of neural radiance
  fields.
\newblock In {\em Proceedings of the IEEE/CVF Conference on Computer Vision and
  Pattern Recognition}, pages 3835--3844, 2022.

\bibitem{clip-nerf}
Can Wang, Menglei Chai, Mingming He, Dongdong Chen, and Jing Liao.
\newblock Clip-nerf: Text-and-image driven manipulation of neural radiance
  fields.
\newblock In {\em Proceedings of the IEEE/CVF Conference on Computer Vision and
  Pattern Recognition}, pages 3835--3844, 2022.

\bibitem{wang2022nerf}
Can Wang, Ruixiang Jiang, Menglei Chai, Mingming He, Dongdong Chen, and Jing
  Liao.
\newblock Nerf-art: Text-driven neural radiance fields stylization.
\newblock {\em arXiv preprint arXiv:2212.08070}, 2022.

\bibitem{wang2023towards}
Haowei Wang, Jiayi Ji, Yiyi Zhou, Yongjian Wu, and Xiaoshuai Sun.
\newblock Towards real-time panoptic narrative grounding by an end-to-end
  grounding network.
\newblock {\em arXiv preprint arXiv:2301.03160}, 2023.

\bibitem{wang2018non}
Xiaolong Wang, Ross Girshick, Abhinav Gupta, and Kaiming He.
\newblock Non-local neural networks.
\newblock In {\em Proceedings of the IEEE conference on computer vision and
  pattern recognition}, pages 7794--7803, 2018.

\bibitem{woo2018cbam}
Sanghyun Woo, Jongchan Park, Joon-Young Lee, and In~So Kweon.
\newblock Cbam: Convolutional block attention module.
\newblock In {\em Proceedings of the European conference on computer vision
  (ECCV)}, pages 3--19, 2018.

\bibitem{wu20153d}
Zhirong Wu, Shuran Song, Aditya Khosla, Fisher Yu, Linguang Zhang, Xiaoou Tang,
  and Jianxiong Xiao.
\newblock 3d shapenets: A deep representation for volumetric shapes.
\newblock In {\em Proceedings of the IEEE conference on computer vision and
  pattern recognition}, pages 1912--1920, 2015.

\bibitem{xu2022dream3d}
Jiale Xu, Xintao Wang, Weihao Cheng, Yan-Pei Cao, Ying Shan, Xiaohu Qie, and
  Shenghua Gao.
\newblock Dream3d: Zero-shot text-to-3d synthesis using 3d shape prior and
  text-to-image diffusion models.
\newblock {\em arXiv preprint arXiv:2212.14704}, 2022.

\bibitem{xu2015show}
Kelvin Xu, Jimmy Ba, Ryan Kiros, Kyunghyun Cho, Aaron Courville, Ruslan
  Salakhudinov, Rich Zemel, and Yoshua Bengio.
\newblock Show, attend and tell: Neural image caption generation with visual
  attention.
\newblock In {\em International conference on machine learning}, pages
  2048--2057. PMLR, 2015.

\bibitem{attngan}
Tao Xu, Pengchuan Zhang, Qiuyuan Huang, Han Zhang, Zhe Gan, Xiaolei Huang, and
  Xiaodong He.
\newblock Attngan: Fine-grained text to image generation with attentional
  generative adversarial networks.
\newblock In {\em Proceedings of the IEEE conference on computer vision and
  pattern recognition}, pages 1316--1324, 2018.

\bibitem{ye2022shifting}
Jiabo Ye, Junfeng Tian, Ming Yan, Xiaoshan Yang, Xuwu Wang, Ji Zhang, Liang He,
  and Xin Lin.
\newblock Shifting more attention to visual backbone: Query-modulated
  refinement networks for end-to-end visual grounding.
\newblock In {\em Proceedings of the IEEE/CVF Conference on Computer Vision and
  Pattern Recognition}, pages 15502--15512, 2022.

\bibitem{KangxueYin20233DStyleNetC3}
Kangxue Yin, Jun Gao, Maria Shugrina, Sameh Khamis, and Sanja Fidler.
\newblock 3dstylenet: Creating 3d shapes with geometric and texture style
  variations.
\newblock {\em international conference on computer vision}, 2023.

\bibitem{stackgan}
Han Zhang, Tao Xu, Hongsheng Li, Shaoting Zhang, Xiaogang Wang, Xiaolei Huang,
  and Dimitris~N Metaxas.
\newblock Stackgan: Text to photo-realistic image synthesis with stacked
  generative adversarial networks.
\newblock In {\em Proceedings of the IEEE international conference on computer
  vision}, pages 5907--5915, 2017.

\bibitem{stackgan++}
Han Zhang, Tao Xu, Hongsheng Li, Shaoting Zhang, Xiaogang Wang, Xiaolei Huang,
  and Dimitris~N Metaxas.
\newblock Stackgan++: Realistic image synthesis with stacked generative
  adversarial networks.
\newblock {\em IEEE transactions on pattern analysis and machine intelligence},
  41(8):1947--1962, 2018.

\bibitem{zhang2022arf}
Kai Zhang, Nick Kolkin, Sai Bi, Fujun Luan, Zexiang Xu, Eli Shechtman, and Noah
  Snavely.
\newblock Arf: Artistic radiance fields.
\newblock In {\em European Conference on Computer Vision}, pages 717--733.
  Springer, 2022.

\bibitem{zhou2016thingi10k}
Qingnan Zhou and Alec Jacobson.
\newblock Thingi10k: A dataset of 10,000 3d-printing models.
\newblock {\em arXiv preprint arXiv:1605.04797}, 2016.

\end{thebibliography}
}

\end{document}